%% file: main.tex
\documentclass[conference]{IEEEtran}

\IEEEoverridecommandlockouts
\usepackage{amsmath,amssymb,amsfonts}
\usepackage{algorithmic} 
\usepackage{graphicx}
\usepackage{array}
\usepackage{mdwmath}
\usepackage{mdwtab}
\usepackage{eqparbox}
\usepackage{url}
\usepackage{hyperref}
\usepackage{tabularx}
\usepackage{subcaption}
\hypersetup{citecolor=Green}
\usepackage{mathrsfs}
\usepackage{rsfso}
\usepackage{multirow}
\usepackage{textcomp}
\usepackage{xcolor}
\usepackage{subfloat}
\usepackage{my_packages}
\usepackage[sorting=none]{biblatex}

\begin{document}
\title{Wind Power Prediction across Different Locations using Deep Domain Adaptive Learning}

\author{
\IEEEauthorblockN{Md Saiful Islam Sajol\IEEEauthorrefmark{1}, Md Shazid Islam\IEEEauthorrefmark{2}, A S M Jahid Hasan \IEEEauthorrefmark{3}, Md Saydur Rahman\IEEEauthorrefmark{2}, Jubair Yusuf}

\IEEEauthorblockA{\IEEEauthorrefmark{1}Louisiana State University, Louisiana, USA, \IEEEauthorrefmark{2}University of California Riverside, California, USA,\\ \IEEEauthorrefmark{3}North South University, Dhaka, Bangladesh,\\ Email: msajol1@lsu.edu, misla048@ucr.edu, jahid.hasan12@northsouth.edu,\\ mrahm054@ucr.edu, jyusuf177@gmail.com}
}


\maketitle

\input{texfiles/0.abstract.tex}
\input{texfiles/1.introduction.tex}

\input{texfiles/2.background_study.tex}

\input{texfiles/3.methodology.tex}

\input{texfiles/4.dataset.tex}

\input{texfiles/5.experiments_and_results.tex}

\input{texfiles/6.conclusion.tex}

\printbibliography

\end{document}

%% file: texfiles/0.abstract.tex
\begin{abstract}
Accurate prediction of wind power is essential for the grid integration of this intermittent renewable source and aiding grid planners in forecasting available wind capacity. Spatial differences lead to discrepancies in climatological data distributions between two geographically dispersed regions, consequently making the prediction task more difficult. Thus, a prediction model that learns from the data of a particular climatic region can suffer from being less robust. A deep neural network (DNN) based domain adaptive approach is proposed to counter this drawback. Effective weather features from a large set of weather parameters are selected using a random forest approach. A pre-trained model from the source domain is utilized to perform the prediction task, assuming no source data is available during target domain prediction. The weights of only the last few layers of the DNN model are updated throughout the task, keeping the rest of the network unchanged, making the model faster compared to the traditional approaches. The proposed approach demonstrates higher accuracy ranging from 6.14\% to even 28.44\% compared to the traditional non-adaptive method. 
\end{abstract}


%% file: texfiles/1.introduction.tex
\section{\textbf{Introduction}}

 A substantial rise in global energy demand is observed due to the growing population, rapid industrialization, and progress in technology. While fossil fuel sources are being depleted to match this increasing demand, they also have adverse effect on our environment. Thus, renewable energy has become an effective solution to this energy crisis. Only in 2023, the renewables installed over the world was 510 GW as reported by the International Energy Agency (IEA) \cite{iea23}. Due to the advancement of technology and lowering costs, wind energy has become one of the prevailing renewable sources. Currently, the global installed wind energy capacity is more than 710 GW \cite{iea23}. This is an increase by a factor of almost 100 times over the past two decades \cite{irena23}.

 Similar to most other renewable resources, wind energy production is highly intermittent and dependent on weather conditions. The fluctuations in wind power can lead to several issues, such as frequency deviation, harmonics, inter-area oscillations, economic dispatch problems within electricity markets, etc. \cite{shakir20}. Therefore, reliable prediction of wind power generation is of colossal importance. However, as wind power is dependent upon weather conditions, the generation varies widely over regions even for similar wind farm setup. Weather variables are location-dependent and deviate vastly from one climatic region to another, consequently causing deviations in wind power generation profile. Wind power generation can be predicted with statistical and machine learning (ML) methods, however, the prediction model remains mostly location-specific. The drawback of location-specific prediction models is that each model needs to be trained individually, requiring a large amount of data. However, reliable data collection in bulk quantities can be a challenging task in some locations due to adverse conditions. In such cases, domain adaptation or transfer learning methods can help immensely. Domain adaptation allows an ML model trained on a source dataset to predict the outcomes of another dataset called the target dataset with a comparable but different probability distribution. Domain adaptation makes the model more robust and computationally more efficient as it becomes more location-agnostic and requires less data from the test dataset. The motivation behind our work revolves around addressing this issue. 

 The contributions of this work can be summarized as:

 \begin{enumerate}
     \item Proposing a domain adaptation-based location-agnostic model to predict wind power generation that achieves both higher accuracy and faster convergence.
     \item Generating a novel dataset that combines features from two distinct datasets and identifying relevant features through the application of random forest. 
 \end{enumerate}

%% file: texfiles/2.background_study.tex
\section{\textbf{Related Works}}

\begin{figure*}
\centering
\includegraphics[width= 0.9\textwidth]{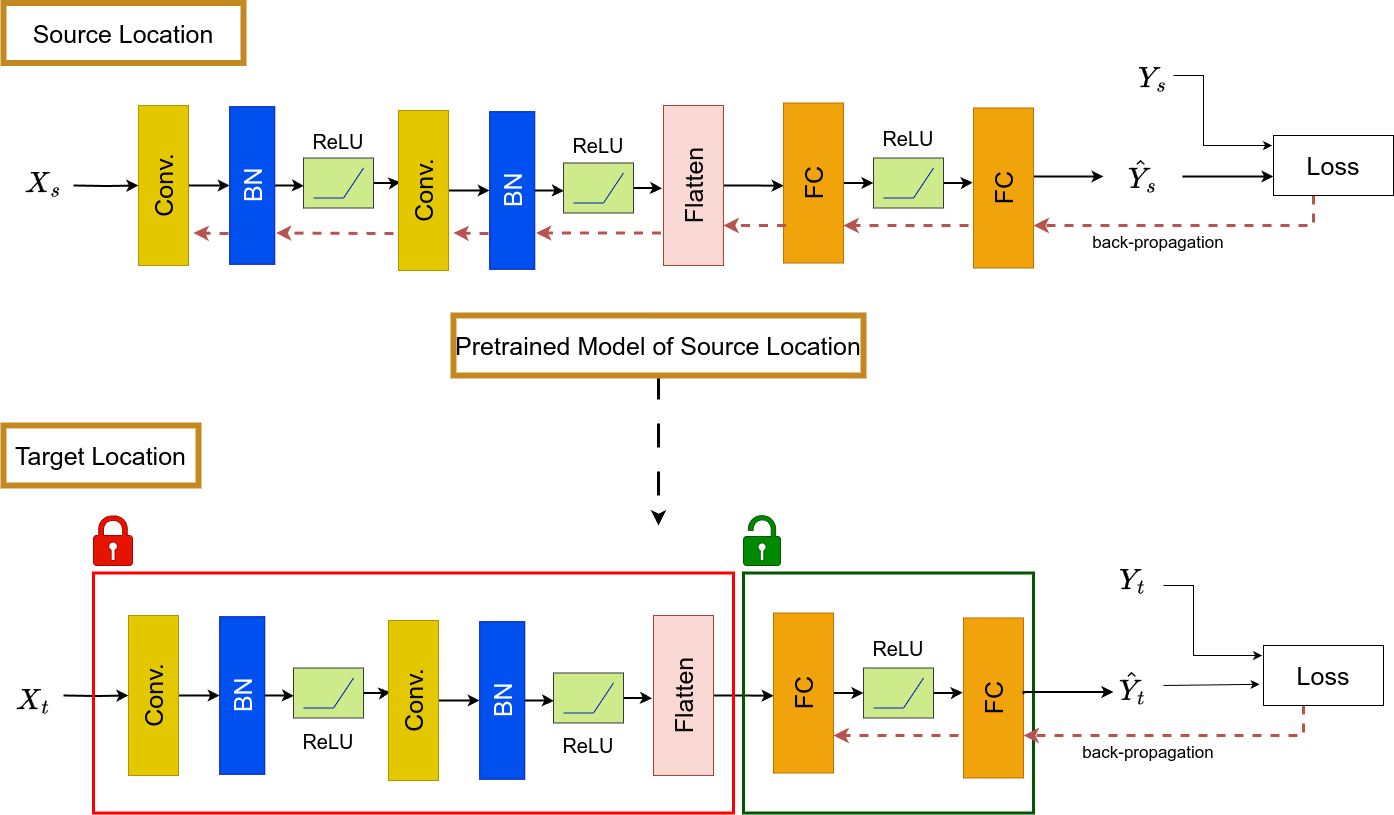}
\caption{Overview of the methodology is shown in the figure. At first, a model is trained from scratch on the source side using the source data $(X_s, Y_s)$. Then the pretrained model of the source side is transferred to the target side. On the target side, the weight of the last two layers (FC layers) of the model is adapted using target data $(X_t, Y_t)$. The rest of the network weights are kept as same as the pre-trained model.\cite{10407265}}
\label{overview}
\end{figure*}

Wind power forecasting has been extensively researched in the literature. Most of the forecasting methods can be broadly classified into three approaches. They are (i) physical approach, (ii) statistical approach, and (iii) machine learning and deep learning-based approach. The physical approach has been investigated for wind power forecasting in several papers \cite{physical2,physical1,physical3}. Statistical methods have also been studied in literature \cite{statistical1, statistical2, statistical3}. However, more recently ML and deep learning-based methods have gained popularity. 

Transfer learning and domain adaptation-based deep learning models have also gained interest in in contemporary literature. Transfer learning is used with auto-encoding to achieve better prediction results for wind forecasting \cite{hu2016, liu2021}.  In \cite{liu2021}, auto encoding and transfer learning have been applied to determine the patterns in homogeneous and heterogeneous characteristics of wind turbine dynamics for wind power prediction. The paper claims that heterogeneous characteristics can stem from differences in operating conditions, locations, and operational strategies, etc. and compares the results of different methods with their proposed method for two wind farms. The proposed method outperforms the other methods in terms of both accuracy and time. However, the reason behind the varying heterogeneous characteristics of these two wind farms is absent in this work. A deep neural network (DNN)-based architecture is proposed with transfer learning application \cite{hu2016}. The goal is to predict the wind speed in newly built wind farms, lacking data, exploiting the data from previously built wind farms, where sufficient data is present. The data from four wind farms in China is used here. However, they are not located far enough to show enough variability in climatic conditions. A novel serio-parallel architecture is proposed in \cite{YIN2021121271}, where several parallel Convolutional Neural Networks (CNN) are separately connected to a Long Short Term Memory (LSTM) network. Transfer learning is applied to this network by adopting the Adam optimization to retrain the frozen layers of the pre-trained feature extractor and the fully connected layers. The dataset is comprised of six separate wind farms although they belong to the same climatic region of Inner Mongolia, China. Model-based transfer learning is applied in \cite{LIU2022118729} to construct a Multi-layer Extreme Learning Machine (MLELM). Output parameters of MLELM are further optimized by using particle swarm optimization. Instead of providing the expected value of wind power, the proposed method provides optimized probability quantiles of the predicted values. However, all eight wind farm data comes from the same province in China. Transfer learning along with gradient boosting method also has been explored for some cases \cite{electronics11244125, en12010159}. An instance-based transfer learning embedded with gradient-boosting decision trees is proposed in \cite{en12010159}. Similar to \cite{LIU2022118729}, this paper also tries to find out the probability quantiles of the predicted wind power. An optimization method is proposed for assigning weights to the auxiliary training sets based on the relatedness of source and target tasks. However, all the ten wind farms used here are from Australia. Two transfer learning approaches Multi-task Learning (MTL) and Bayesian Embedded Multi-task Learning (BE-MTL) are proposed and compared with other regular ML methods namely, Random Forests, Gradient Boosted Regression Trees, and simple Multi-Layer Perceptrons \cite{vogt2019wind}. The MTL and BE-MTL methods show superior performance while working with less available data. However, all 19 wind farm data are from southwest Germany. In \cite{tasnim2018wind}, instead of using data from a single wind farm as a source domain, a cluster of wind farm data is used as multiple source domains. This multi-source domain adaptation (MSDA) uses different weights for each source domain with different but comparable probability distributions. This method shows a significant reduction in prediction error. Nonetheless, the work focuses on wind farms with similar weather patterns and thus uses wind stations that are closely located.   Recently \cite{10407265,10456479} has shown the application of domain adaptation in solar power and rain precipitation prediction, respectively in a location-agnostic manner. 

As we can see from above, there are multiple research works that focused on the use of transfer learning or domain adaptation to improve the forecasting results of wind power. Yet these works have not leveraged the fact that a pre-trained model of one climatic region's wind farm can be used to predict the wind power generation of other climatic regions through domain adaptation. Therefore, our work centers around this issue and tries to generate a location-agnostic domain adaptive model.

%% file: texfiles/3.methodology.tex
\section{\textbf{Methodology}}
 In this paper, we propose a method to predict the wind power generation in a region at a certain time, based on the weather and climatological characteristics of the region. A model that was once trained using the data of one country will be used further to adapt to the domain of a different country. The first country where the model was actually trained is referred as the source domain, and the other country where we intend to adapt the model is referred as the target domain. In the source domain, the model is trained in a fully supervised manner. For this work, we assume that the target domain does not have any data available, i.e. the only thing it has is a pre-trained model from the source domain. Then the pre-trained model is adapted for the target domain by only updating the last few layers of the pre-trained model. Therefore, the overall training procedure becomes faster and the computational burden is reduced for larger datasets. In short, our work follows two main steps, i.e., train the model in the source domain, followed by adapting the model in the target domain. Figure \ref{overview} illustrates the overview of the overall approach.

\subsection{Training the model in source domain’s wind dataset}

We defined our prediction task as a classification problem. As the wind power generated in a country varies over time, to construct it as a classification problem, the values are labeled with certain classes depending on the range where it lies. We divided the range of wind power generation into N number of bins and labeled the rows with their corresponding classes. Figure \ref{hist} shows the histogram of the generated wind power of the source country. To sum up, we have categorized wind power generation into several classes and represented this as a classification problem within the framework of deep learning. To evaluate our model's performance in predicting the correct binned values of wind power, we have chosen accuracy as the primary performance metric.
Here the accuracy is defined as the proportion of true results (both true positives and true negatives) among the total number of cases examined. In the script, accuracy is determined by comparing the model's predicted labels with the actual labels after making predictions. 

In the source domain, the model is trained with a two layer deep convolutional network strategically designed to capture both local and global patterns of the weather features from the given input data. The DNN is comprised of two convolutional layers (Conv.), two batch normalization layers (BN), two fully connected layers (FC), and the rectified linear unit (ReLU) layers. The initial convolutional layers are used to extract features and to systematically analyze spatial patterns in the wind data. Batch normalization and ReLU activation functions stabilize the network's training process and enhance its robustness. The ReLU layer also helps to learn intricate relationships and patterns present in the wind data. The output FC layer has an N number of nodes which is equal to the number of bins N made by using the categories for the dataset. Cross Entropy loss CE, the network is then trained for the Xs input data. Convergence is monitored by tracking the reduction in the loss function CE during training epochs. When significant improvements in the model's performance cease, it implies that the model has maximized its learning from the training data. Cross Entropy loss achieves superior results by handling probabilistic outputs and adapting to the class imbalance in comparison to other traditional loss functions.

\subsection{Adaptation in Target Domain’s Dataset}

On the target side, we assume the source data are not available and we have only a DNN model pre-trained in the source country. The weights of the pre-trained model will be updated so that it performs better for the target domain. Though it is possible to update all the weights of the model, it is computationally expensive. So to reduce the computational cost, the entire model's weights are kept frozen except the last two FC layers. This mechanism helps model to learn faster without any significant changes in weights inside its entire model.



%% file: texfiles/4.dataset.tex
\begin{figure}
\centering
\includegraphics[width=\columnwidth]{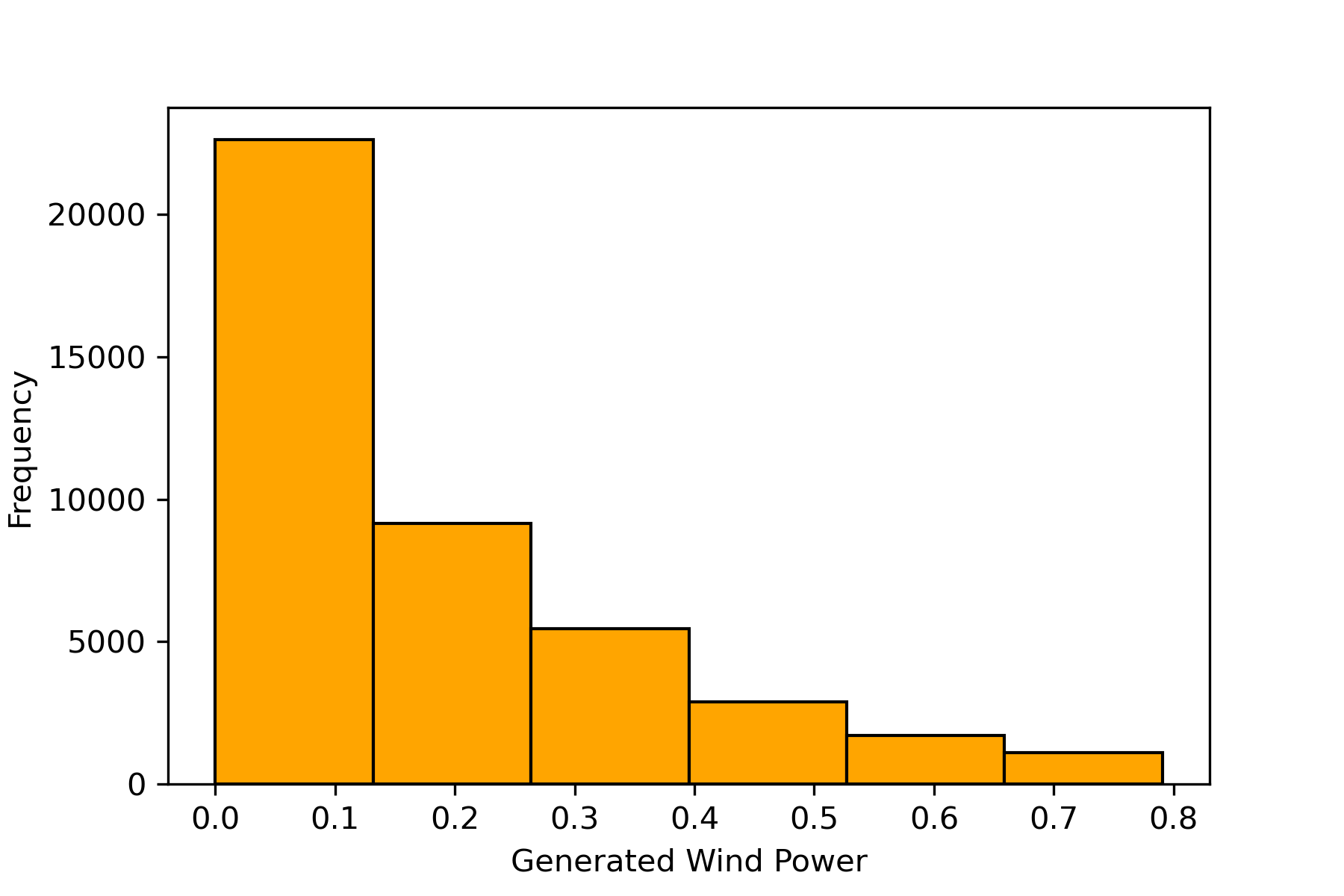}
\caption{Histogram of wind power generation shown in 6 bins. Note, that the values along the x-axis are normalized by their maximum generated power.}
\label{hist}
\end{figure}
\begin{figure}
\centering
\includegraphics[width=\columnwidth]{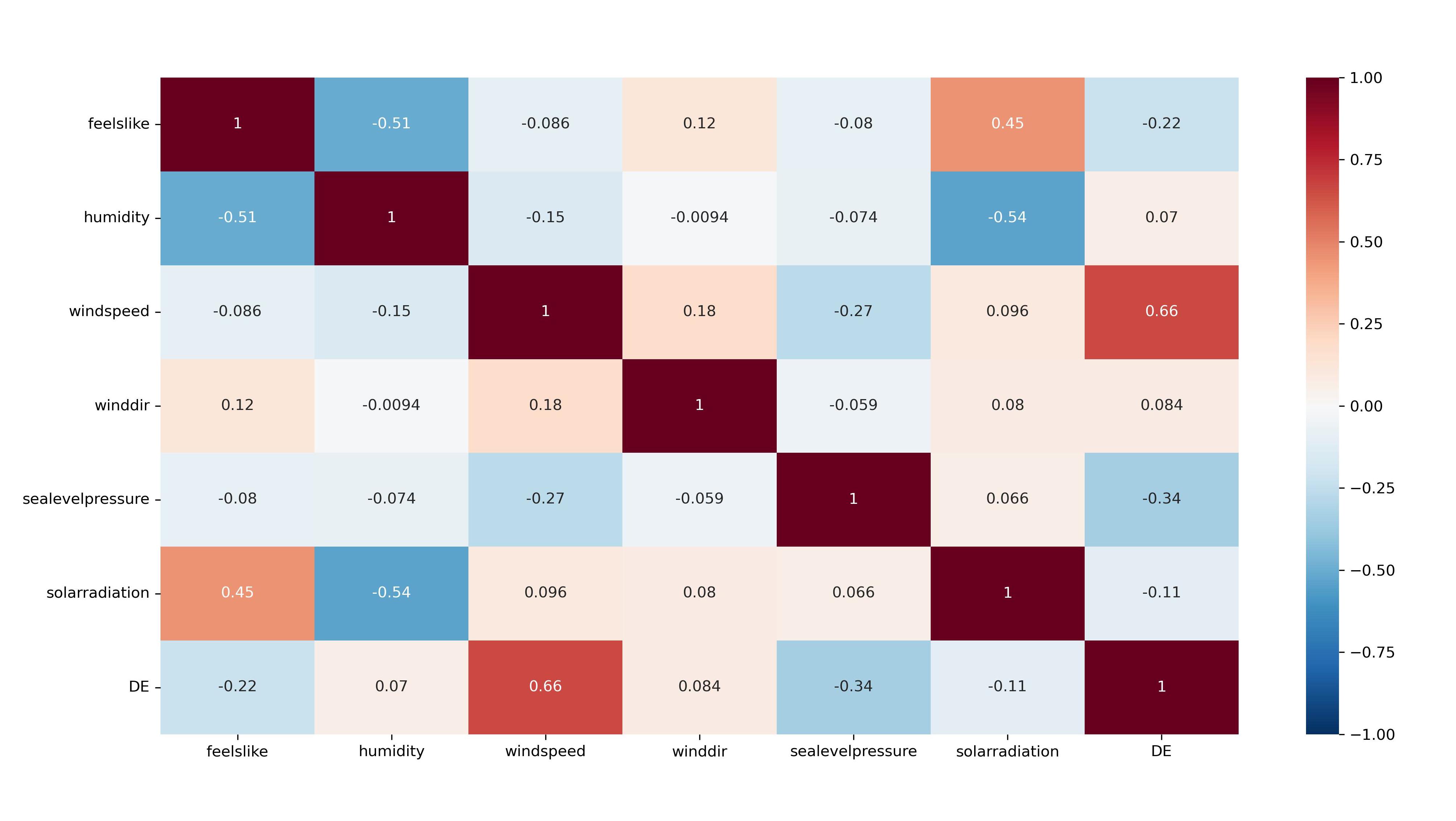}
\caption{Correlation matrix of the selected features}
\label{correlation}
\end{figure}
\section{\textbf{Dataset}}

\begin{table}[h]
\centering
\caption{Dataset Features and Data Types}
\label{tab:dataset_features}
\begin{tabular}{|l|l|l|l|}
\hline
\textbf{Feature}       & \textbf{Data Type} & \textbf{Feature}       & \textbf{Data Type} \\ \hline
temp                   & Float              & winddir                & Integer \\ \hline
feelslike              & Float              & sealevelpressure       & Float \\ \hline
dew                    & Float              & cloudcover             & Float \\ \hline
humidity               & Float              & visibility             & Float \\ \hline
precip                 & Float              & solarradiation         & Float \\ \hline
precipprob             & Float              & solarenergy            & Float \\ \hline
preciptype             & String             & uvindex                & Integer \\ \hline
snow                   & Float              & severerisk             & Integer \\ \hline
snowdepth              & Float              & conditions             & String \\ \hline
windgust               & Float              & icon                   & String \\ \hline
windspeed              & Float              &                        &        \\ \hline
\end{tabular}
\end{table}

\begin{table*}[!h]
  \caption{Comparing the performance in terms of percentage accuracies for Deep Learning Network on different locations before and after adaptation. }
  \small\centering
 
    \begin{tabular}
    {|c| c c c| c c c| c c c|}
    \hline
    & \multicolumn{9}{c|}{ \textbf{Target Domain} } \\
    \cline { 2 - 10 } 
    \textbf{Source} &  \multicolumn{3}{c|}{\textbf{\textit{ Germany}} } & \multicolumn{3}{c|}{\textbf{\textit{France}}} & \multicolumn{3}{c|}{\textbf{\textit{UK}}} \\
    \cline { 2 - 10 }
    \textbf{Domain} & \textbf{\textit{w/o adapt.}} & \textbf{\textit{w. adapt.}} & \textbf{\textit{diff.}} & \textbf{\textit{w/o adapt.}} & \textbf{\textit{w. adapt.}} & \textbf{\textit{diff.}}  & \textbf{\textit{w/o adapt.}} & \textbf{\textit{w. adapt.}} & \textbf{\textit{diff.}} \\
    \hline $\mathrm{Germany}$ & \multicolumn{3}{c|}{$\mathrm{N} / \mathrm{A}$} & 53.25 $\%$ & 67.25 $\%$ & 14.00 $\%$ & 43.19 $\%$ & 58.12 $\%$ & 14.93 $\%$\\
    \hline $\mathrm{France}$ & 65.18 $\%$ & 71.32  $\%$ & 6.14 $\%$ & \multicolumn{3}{c|}{$\mathrm{N} / \mathrm{A}$} & 43.02 $\%$ & 58.83 $\%$ & 15.80 $\%$\\
    \hline UK & 42.12 $\%$ & 70.56 $\%$ & 28.44 $\%$ & 43.93 $\%$ & 67.42 $\%$ & 23.49 $\%$ & \multicolumn{3}{c|}{ N/A } \\
    \hline
    \end{tabular}

\label{table:adapt}
\end{table*}

An original dataset is created for domain adaptive wind power prediction. The creation of this novel dataset is executed through incorporation of appropriate data from two different datasets; one containing the hourly wind generation data and another containing the meteorological data. The wind generation data is collected from the EMHIRES dataset \cite{emhires}. This dataset contains 30 years' worth of hourly wind generation data for 36 of the European countries. For each country, the wind generation is normalized relative to the installed wind generation capacity. For this work, the latest available five years' worth of data has been used. Hourly meteorological data is retrieved from \cite{viscros} for the corresponding wind power generation period. The final dataset is created by merging the two datasets according to their corresponding hourly slots. Ultimately, the meteorological dataset contains 21 weather parameters shown in Table~\ref{tab:dataset_features}. However, only a handful of the weather features are useful for wind power prediction as inclusion of irrelevant features can make the training process slower and may even compromise the fidelity of the forecast model.

Random forest algorithm is considered as an effective tool for the selection of compatible features \cite{randfor}. Following the implementation of this algorithm, only six effective features were selected for prediction namely, temperature, dew point, snow, snow depth, wind speed, and cloud cover.From the dataset, we have analyzed the correlation of data for wind power generation.  Figure \ref{correlation} displays the correlation among the selected six features related to wind power generation through a heat map. For our experiments, we choose three different countries Germany, France, and the United Kingdom. The location of these three countries is such that they have a comparable yet distinguishable weather profile. The dataset comprises wind data and various weather features from three distinct European countries, effectively capturing the spatial disparities in data distribution across these regions. However, while the dataset acknowledges these differences, it's our proposed model that solves the challenge of properly addressing them.

%% file: texfiles/5.experiments_and_results.tex
\section{\textbf{Experiments and Results}}

\subsection{Experimental Setup}
 The implementation of all the experiments is executed by the popular deep learning framework Pytorch \cite{pytorch}. A learning rate of 0.001, batch size of 64, and a well-known optimization technique Adam optimizer were used for training.





    
    



    
    

\subsection{Performance of Adaptation on Accuracy:}

 The model from the source domain (country) was tested to predict wind power in the target domain (another country). The pre-trained model is further trained to adapt the performance to the target domain. To check the domain adaptation extensively, in this experiment, each pre-trained model was adapted for the rest of the domains. In TABLE \ref{table:adapt}, we demonstrate how the results improve after adaptation in terms of accuracy. The table presents that, the performance of Germany (target domain) improves by 6.14\% and 28.44\% when they were adapted from the source domain, France and the UK, respectively. France being a target domain shows an improvement of 14.00\% and 23.49\% when adapted from Germany and the UK, respectively. Lastly, when the UK was used as the target domain the accuracy jumped by 14.93\% and 15.80\% for the source domain of Germany and France, respectively. Overall, in all the cases, the performance of the models improved significantly after domain adaptation. The first few layers of the model learn the general feature distribution, while the final layers learn to specialize the specific task to be performed. As our model only updates the last few layers, it allows us to retain the generalized knowledge of the source task and make special adjustments to the target task through fine-tuning, achieving higher accuracy.

\begin{figure*}
\centering

    \includegraphics[width= \textwidth]{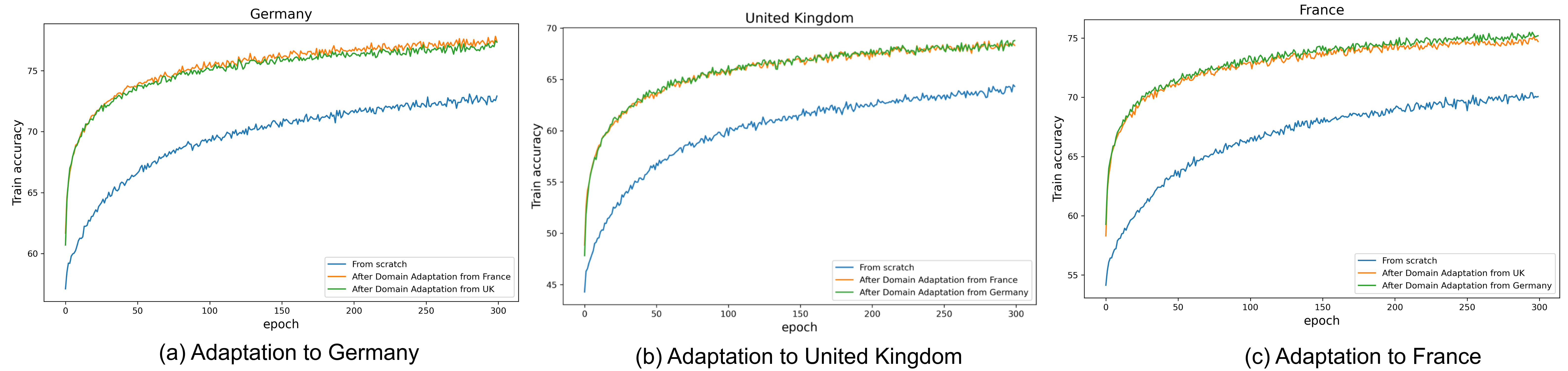}
    \caption{Comparing the temporal performance between training from scratch and using adaptation from a pre-trained model for three countries - Germany, United Kingdom and France. The graphs show that with domain adaptation, the network completes the training (reaches the saturation of accuracy) faster.}

    \label{time}
    
\end{figure*}

\subsection{Performance comparison between training from scratch and using adaptation :}

Fig. \ref{time} shows the temporal performance comparison between training a model from scratch and our proposed adaptation technique.In these figures, the lines representing performance after adaptation are higher than the "from scratch" line. This indicates that adapting the model for the newer domain improved its performance compared to training it solely on a different source domain data. Additionally, we observe that using the adaptation technique allows the network to reach saturation in less time which reduces the computation cost. The same patterns are observed for all three countries - Germany, the United Kingdom, and France. That means in all the cases the initial accuracy of the models is improved, learned at a faster rate than the model developed from scratch, and the final accuracy is also higher than a model from scratch.


%% file: texfiles/6.conclusion.tex
\section{\textbf{Aditional Studies}}
\subsection{Performance comparison between updating full network and partial network:}
In this work, only the last two
layers (FC layers) of the network are updated while working on the target
domain rather than updating the whole network. In TABLE
II, the performance between updating the full
network and the partial network is compared considering Germany, as the source
domain and France and, the UK as the target domain. The table additionally takes into account scenarios where France and the UK serve as the source domains, with the remaining countries acting as target domains. It is found that updating the whole network during adaptation increases
accuracy minimum of 0.04\% to a maximum of 1.45\%. However, updating the full network increases the computational cost as it requires the entire network to further train. The minimal effect on accuracy compared to the considerable decrease in computation time directed us to use the partial update strategy when using domain adaptation techniques for a new target region.



\newcolumntype{C}[1]{>{\centering\arraybackslash}p{#1}}


\begin{table}[h]
\centering
\caption{Comparative accuracy (\%) metrics showing the impact of updating the full network vs updating partial network}
\label{tab:performance_metrics}
\begin{tabular}{|C{0.85cm}|C{1cm}|C{1.9cm}|C{1.9 cm}|C{1.1cm}|}
\hline
\textbf{Source} & \textbf{Target} & \textbf{Updating with Partial Network} & \textbf{Updating with Full Network} & \textbf{Difference} \\
\hline

\multirow{2}{*}{Germany} & France  & 67.25 & 68.56  & 1.31 \\  
                         & UK      & 59.7 & 60.85  & 1.15 \\
                         
\hline
\multirow{2}{*}{France}  & Germany & 71.32 & 71.36  & 0.04 \\  
                         & UK      & 58.83 & 60.28  & 1.45 \\ 
                         
\hline
\multirow{2}{*}{UK}      & Germany & 70.56 & 70.74  & 0.18 \\ 
                         & France  & 67.42 & 67.88  & 0.46 \\ 
                         
\hline
\end{tabular}
\end{table}

\subsection{Performance comparison between with and without ef-
fective feature selection:}

 Out of the 21 features available in the dataset, only six effective features were selected for training the model via the application of random forest. Table III presents the outcomes of an ablation study. This study investigates the effect of feature selection by comparing the performance of the model with utilizing all features against considering selected features only. As demonstrated in the table, employing all features results in a maximum accuracy increase of only 2.54\% for Germany.  This increment is achieved at the cost of training fifteen additional features.  Whereas, training with just six features doesn't significantly compromise model accuracy, offering a considerably more computationally efficient training process.
 
\begin{table}
    \centering
    \caption{Comparative accuracy (\%) metrics showing the impact of using all features vs using selected features only}
    \begin{tabular}{|c|c|c|c|} \hline 
         Location&  Taking All Features& Taking selected Features & Difference\\ \hline 
         France&  67.32& 65.33 & 1.99  \\  \hline 
         Germany&  71.50& 68.96 & 2.54 \\ \hline 
         UK&  60.23& 59.38 & 0.84 \\ \hline
    \end{tabular}
    \label{tab:my_label}
\end{table}

\section{\textbf{Conclusion \& Future Work}}

In this work a source-free and location-agnostic domain adaptive model was presented for wind power generation prediction. The proposed model gets pre-trained on one country's data (source domain) and this model is then used to predict the wind generation of another country (target domain) using selective weather features. The proposed approach performs significantly better than the traditional approach. The assumption of source data being unavailable while training the target domain makes it less data-intensive and computationally more efficient. Furthermore, two additional studies are carried out to investigate the impact of updating the partial network during target domain training and the impact of using selective features only, exhibiting the efficacy of our design. While our current implementation leverages supervised learning, we plan to explore unsupervised learning techniques in the future for further extension and refinement of our approach. Additionally, we aim to incorporate federated learning methodologies, bolstered by robust security features, to enhance the privacy and integrity of our model across distributed environments.